\documentclass[runningheads]{llncs}
\usepackage[T1]{fontenc}
\usepackage{amsmath,amssymb,amsfonts}
\usepackage{cite}
\usepackage{times}
\usepackage{algorithm,algorithmic}
\usepackage{subfigure}
\usepackage[pdftex]{graphicx}
\usepackage{textcomp}
\usepackage{color}
\usepackage{xcolor}
\usepackage{pgfplots}
\pgfplotsset{compat=1.16}
\usepackage{tikz}
\usepackage{bm}
\usepackage{hyperref} 
\usepackage{soul}
\usepackage{booktabs}
\usepackage{multirow}
\usepackage{graphicx,lipsum}
\usepackage{todonotes}
\usepackage{floatpag}
\usepackage{xspace}
\newboolean{useComments}

\setboolean{useComments}{false}

\definecolor{dark_purple}{rgb}{0.1, 0.0, 0.4}
\definecolor{dark_green}{rgb}{0.0,0.2,0.5}
\definecolor{dark_red}{rgb}{0.85,0, 0}

\ifthenelse{\boolean{useComments}}{%
	\newcommand{\mh}[1]{\todo[inline,color=white!40,bordercolor=white]{\textcolor{teal!70!black!80}{\textbf{Moritz:}\textmd{\;#1}}}} 
	\newcommand{\ad}[1]{\todo[inline,color=white!40,bordercolor=white]{\textcolor{red}{\textbf{Akshay:}\textmd{\;#1}}}}
	\newcommand{\vh}[1]{\todo[inline,color=white!40,bordercolor=white]{\textcolor{purple}{\textbf{Vahid:}\textmd{\;#1}}}}
	\newcommand{\ns}[1]{\todo[inline,color=white!40,bordercolor=white]{\textcolor{dark_purple}{\textbf{Nicolas:}\textmd{\;#1}}}}     
}{%
	\newcommand{\mh}[1]{}
	\newcommand{\ad}[1]{}
	\newcommand{\vh}[1]{}
	\newcommand{\ns}[1]{}
}

\DeclareMathOperator{\newRobust}{\rho}
\DeclareMathOperator{\traj}{\bm{\xi}}
\DeclareMathOperator{\modelParams}{\bm{\theta}}
\DeclareMathOperator{\optimParams}{\bm{\delta}}

\begin{document}
\title{Autonomous Vehicles Path Planning under Temporal Logic Specifications}
%
%
%
\author{Akshay Dhonthi\inst{1,2} \and
Nicolas Schischka\inst{1,3} \and
Ernst Moritz Hahn\inst{2} \and
Vahid Hashemi\inst{1}}

\authorrunning{A. Dhonthi et al.}
%
\institute{AUDI AG, Auto-Union-Stra\ss e 1, 85057, Ingolstadt, Germany \and
Formal Methods and Tools, University of Twente, Enschede, Netherlands \and
Technical University of Munich, Germany\\}

%
\maketitle              
\begin{abstract}
Path planning is an essential component of autonomous driving. 
A global planner is responsible for the high-level planning.
It basically performs a shortest-path search on a known map, thereby defining waypoints used to control the local (low-level) planner.
Local planning is a runtime verification method which is repeatedly run on the vehicle itself in real-time, so as to find the optimal short-horizon path which leads to the desired waypoint in a way which is both efficient and safe.
The challenge is that the local planner has to take into account repeatedly incoming updates about the  information available of the environment.
In addition, it performs a complex task, as it has to take into account a large variety of requirements, originating from the necessity of collision avoidance with obstacles, respecting traffic rules, sticking to regulatory requirements, and lastly to reach the next waypoint efficiently.
In this paper, we describe a logic-based specification mechanism which fulfills all these requirements.


\end{abstract}

\section{Introduction}
\vspace{-0.40cm}

\vh{Please discuss in abstract, conclusion and introduction why this work is related to RV.}
Autonomous driving has gained importance in the recent years.
Path planning is one of the key aspects of automatically steering a driving vehicle.
Here, a trajectory is generated between the current position and the goal position. 
A classical path planner consists of a global (high-level) planner and a local (low-level) planner.
Before the start of a travel, the global planner must find a path to the final goal and generate a full route based on the environment map, thereby defining waypoints which it forwards to the local (low-level) planner.
Global planning is typically defined as a reachability problem to find a path to the goal state.
On the other hand, the local planner is a runtime validation entity which repeatedly and in real-time plans the next few seconds of a trajectory based on both static and dynamic obstacles.
Learning from Demonstrations (LfD) is one of the path planning techniques, in which a demonstrator manually moves the vehicle from start to goal state.
These demonstrations are then learned by the vehicle to reproduce a new path that is close to the demos.
Most of the time, the reproduced trajectory fails to achieve the goal without violating road rules and hitting obstacles. 
For both planners, it is essential to verify during the runtime of the travel that a safe trajectory is generated.

\begin{figure}[t]
	\includegraphics[width=\textwidth]{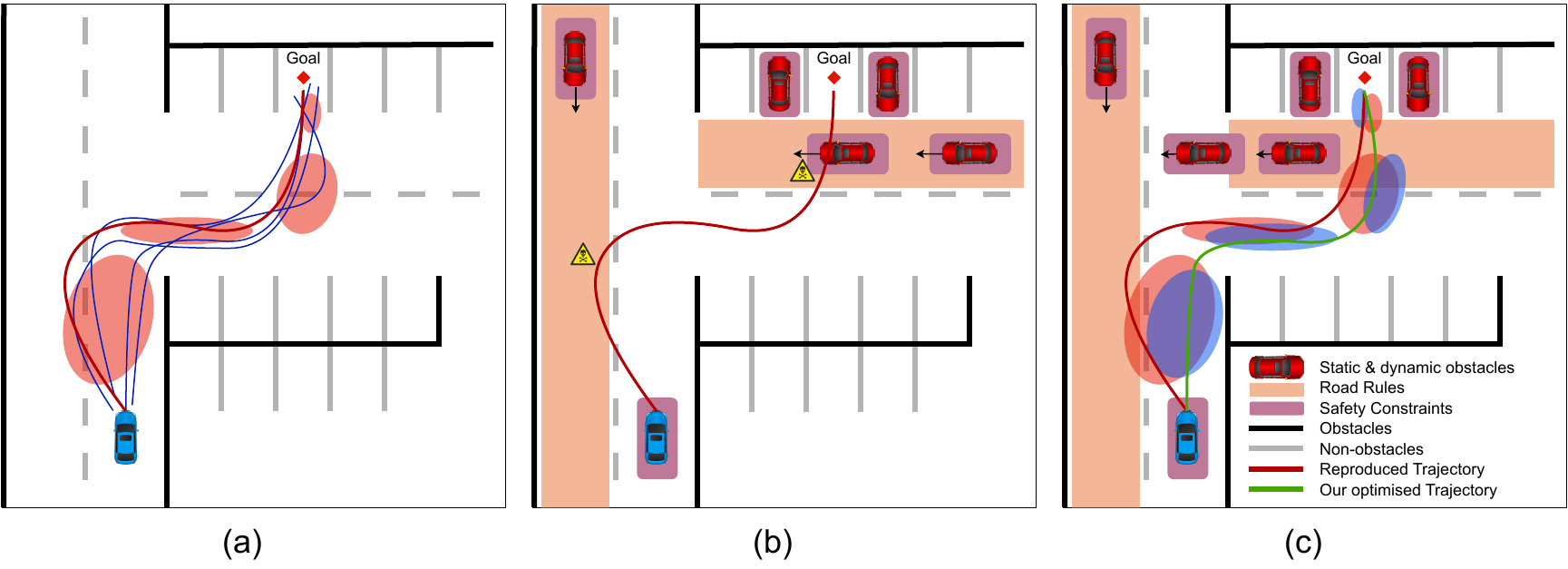}
	\caption{Illustration of the approach. (a) Collection of human demos (in blue) and corresponding spatial GMM (red ellipses). Reproduced trajectory is in red. (b) Real-time scenario during runtime with three different constraint categories labelled in figure, the constraints breach is represented as yellow danger sign. (c) Optimized trajectory (in green) after running our algorithm.
 \mh{again, must make more clear how these ellipses relate to the trajectories generated. That is, if I understood correctly, that trajectories must pass through these ellipses.}
 }
	\label{Fig: Framework}
	\vspace{-0.50cm}
\end{figure}

The LfD method \cite{pignat2017learning} that we use in this work encodes demonstrations using a discrete-state Hidden semi-Markov Model (HSMM) \cite{murphy2002hidden}.
The states of this HSMM are then used as desired waypoints to generate the trajectories using Gaussian Mixture Models (GMM) \cite{calinon2016stochastic} as depicted in Fig.~\ref{Fig: Framework}~(a).
LfD for path planning, however, suffers from limitations to address safety concerns.
Namely, reproduced paths must adhere to road-rules, for example to avoid crossing into the opposite lanes when not necessary.
There can be other obstacles such as cars, bicycles, or pedestrians that might move into the planned trajectory. 
These obstacles can be static or dynamic.
We depict these safety concerns as an example in Fig.~\ref{Fig: Framework}~(b) where we show how the reproduced trajectory fails to adhere to some of the safety concerns mentioned above.

To account for the latter and to address the aforementioned limitations, we propose a new method which will optimize the learned parameters of the LfD (also called model parameters) and generate optimal trajectories.
We formalize safety properties using the logical specification called Signal Temporal Logics (STL)~\cite{mehdipour2019arithmetic}. 
Safety properties could essentially also include requirements from relevant standards such as ISO 26262~\cite{palin2011iso} or SOTIF~\cite{pimentel2019safety}, and we use the logic to specify the requirements of these standards as STL formulas.
Other potential safety properties are also driven by obstacles and road-rules.
Synthesizing optimal model parameters needs a reward function which is a quantitative semantics calculated from the given STL property.  
We utilize the algorithm from our previous work \cite{dhonthi2022optimizing} to achieve this. 
In Fig.~\ref{Fig: Framework}~(c), we illustrate the resultant optimal trajectory (in green) and optimized spatial GMMs (blue ellipses).
As we can see, the new trajectory adheres to all the safety properties.
In Fig.~\ref{Fig: Framework}~(c), we highlight the dynamicity of the obstacles by presenting them at a later time step. 
Our synthesized optimal trajectory can successfully avoid hitting those moving obstacles.

Similar to our work, there are few methods that define safety constraints via linear temporal logic (LTL) in the context of path planning \cite{fainekos2005temporal, lacerda2019probabilistic, rizaldi2018formally}. 
Similar to our work, Barbosa et al. \cite{barbosa2021formal} use STL to incorporate different kinds of constraints and use them in a path planner algorithm called Rapid Exploring Random Tree* (RRT*) \cite{karaman2011sampling}. 
However, these methods do not use LfD for path planning and therefore cannot be scaled up for complex scenarios.  
GMR-RRT* \cite{wang2022gmr} is one approach that is as well close to our work.
This method also learns GMMs to fit human demonstrations but applies a Gaussian mixture regression on the demos unlike ours where we apply HSMM.
The trajectory reproduction is based on a sampling process via the RRT* algorithm \cite{karaman2011sampling} and therefore does not output smooth trajectories. 
The core difference to this approach is that the reproduced paths of \cite{karaman2011sampling} do not account for safety constraints and dynamic obstacles.

\begin{figure}[t]
	\includegraphics[width=\textwidth]{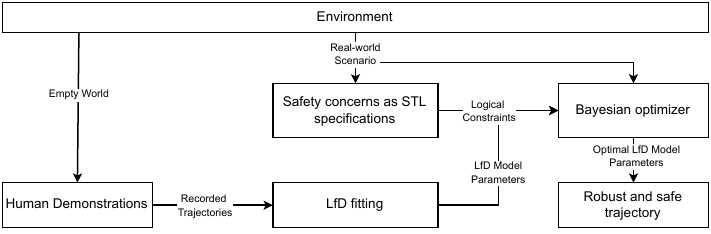}
	\caption{Illustration of the approach.}
	\label{Fig: Flowchart}
	\vspace{-0.80cm}
\end{figure}

Our approach, illustrated in Fig.~\ref{Fig: Flowchart}, starts by collecting human demos from multiple start states and a single goal state. 
Using the trajectories from the demos, we fit an LfD model using the HSMM approach which learns model parameters that match the recorded demos best.\mh{what does that mean, "and obtain model properties"?}\ad{rephrased above sentence, please check, thank you}
Afterward, we define safety constraints based on real-world observations in the form of STL specifications.
In the next step, we run a Bayesian optimizer to optimize the parameters of the LfD model so as to maximize the robustness degree computed using the defined STL specifications.
Finally, we run the optimal trajectory obtained from the optimized LfD model parameters in a real-world environment.

Overall, our contributions in this paper are as follows:
\vspace{-0.20cm}
\nolinebreak
\begin{itemize}
    \item We evaluate the algorithm from our previous work~\cite{dhonthi2022optimizing} on a path planning use case.
    \item We propose a method for continuous path planning to use as a local planner.
    \item We define static and dynamic obstacles as temporal logic constraints and propose a new method to compute robustness for dynamic constraints.
    \item We evaluate the method on two scenarios of an automated valet parking use case.
\end{itemize}

\vspace{-0.50cm}
\section{Preliminaries}
\vspace{-0.30cm}

\subsection{Learning from Demonstrations}
\vspace{-0.20cm}
In this section, we explain the HSMM-based LfD technique from \cite{murphy2002hidden} and how we utilize it in our approach.
We first manually move the ego vehicle (the vehicle we are synthesizing the trajectory for) from the initial state to the goal state and record $N$ demonstrations in the form of trajectories $\traj = \{\traj^i\}_{i=1}^N$, where $\traj^i = \{\traj_t^i\}_{t=1}^T$ with $\traj_t^i \in \mathcal{X} \subseteq \mathbb{R}^{m}$ is the state of the system in $m$ dimensions.
The $m$ dimensions can be vehicle position, velocity, steering angle, etc.
Each $\traj^i$ records spatial co-ordinates and orientation angle $(x, y, \alpha)$  at each time step $t \in {1,\dots,T}$ as we utilize a non-holonomic kinematic model of a differential drive vehicle~\cite{klancar2017wheeled}.
Note that the HSMM model we are using is not only restricted to the automotive area but can also easily be adapted to a different application.

Next, the recorded demonstrations $\traj_t$ are associated\mh{what does that mean, associated with?} with a discrete hidden state sequence $\{z_t\}_{t=1}^T$ with $z_t \in \{1,\dots,K\}$, where $K$ defines the number of components.
Each component represents a specific segment of the trajectory (depicted as red ellipses in Fig.~\ref{Fig: Framework}~(a)).
To move from one segment $i$ to another $j$, a transition matrix $\bm{a} \in \mathbb{R}^{K \times K}$ is learned with $a_{i,j} = P(z_t = j | z_{t-1} = i)$. 
For the next state $j$, we fit multivariate Gaussian distributions written as $\{\bm{\mu_j}, \bm{\Sigma_j}\}$ that represent the demonstrations $\traj_t$.
The parameters $\{\mu_j^S, \Sigma_j^S\}$ denote the duration to stay in a state $j$ for $s$ consecutive steps; we learn their values by fitting a Gaussian $\mathcal{N}(s|\mu_j^S, \Sigma_j^S)$. 
We define the parameter space as $\modelParams = \{ \{a_{i,m}\}_{m-1}^K, \bm{\mu_i}, \bm{\Sigma_i}, \mu_i^S, \Sigma_i^S \}_{i=1}^K$.
We refer the readers to \cite{tanwani2020generalizing} for more details about the approach. 
Since we are solving a non-linear system model, we replace the so-called linear quadratic tracker for the trajectory generation with an iterative linear quadratic regulator~\cite{tassa2012synthesis}.
Using an expectation maximization algorithm, we then train these parameters using the likely state sequence $\bm{z_t} = \{z_1,\dots,z_T\}$.

After learning $\modelParams$ using the demonstrations, we can reproduce a deterministic trajectory $\traj_t'$ (in red) as depicted in Fig.~\ref{Fig: Framework}~(a) (cf.~\cite{tanwani2020generalizing}).
We define $\optimParams \subset \modelParams$ to be the parameters to optimize, where $\optimParams = \{\{a_{i,m}\}_{m-1}^K, \bm{\mu_i}, \mu_i^S\}_{i=1}^K$.
The parameters $\{\bm{\mu_i}\}_{i=1}^K$ represent the spatial position of the Gaussian for HSMM state $i$; changing them will translate the Gaussian in $(x, y)$ directions.
These parameters are useful to correct the trajectory from going into the opposite lane, or to maintain a safe distance to obstacles.
The parameters $\{\mu_i^S\}_{i=1}^K$ represent the temporal state for staying inside an HSMM state $i$ and changing it will reduce or increase the time spent in a region.
This parameter is useful to avoid hitting a moving obstacle by increasing the time spent in the previous state.
Finally, the parameters $\{\{a_{i,m}\}_{m-1}^K\}_{i=1}^K$ represent the sequence in which each HSMM state has to be visited.
It is useful to skip an HSMM state if it is not necessary anymore to satisfy the defined properties.
In this work, we optimize parameters $\optimParams$ to obtain $\hat{\modelParams}$ which can in turn reproduce the trajectory $\hat{\traj}_t$ (in green in Fig.~\ref{Fig: Framework}~(c)) that satisfies the set temporal logical constraints.

\vspace{-0.40cm}
\subsection{STL Specifications}
We recursively define STL formulas according to the following grammar:
\begin{equation}
    \begin{aligned}
        \varphi \;  := & \; \pi^\mu \; | \; \neg\varphi \; | \; \varphi_1\land\varphi_2 \; | \; \mathbf{F}_{[a,b]}\varphi \; | \; \varphi_1 \mathbf{U}_{[a,b]}\varphi_2 ,
    \end{aligned}
    \label{Eqn: STL}
\end{equation}
where $\varphi_1, \varphi_2$ are recursively defined STL formulas, $\pi^\mu\colon \mathcal{X} \rightarrow \mathbb{B}$ is an atomic predicate, the sign of a function $\mu\colon \mathcal{X} \rightarrow \mathbb{R}$ determines whether $\pi^\mu$ is true or false.
By $\traj \models \varphi$ we denote that the demonstration $\traj$ satisfies the STL formula $\varphi$.
Therefore, $\traj \models \mathbf{F}_{[a,b]} \varphi$ iff $\varphi$ holds at some time step between $[a, b]$. 
Similarly, $\traj \models \varphi_1 \mathbf{U}_{[a,b]}\varphi_2$, iff $\varphi_1$ holds until $\varphi_2$ eventually holds during a time step within $[a, b]$.
We can then define the \emph{globally} operator $\mathbf{G}_{[a,b]}\varphi = \neg\mathbf{F}_{[a,b]}(\neg\varphi)$, meaning, $\traj \models \mathbf{G}_{[a,b]} \varphi$ holds within $[a, b]$.

The robustness degree or quantitative semantics for STL denoted as $r(\pi^\mu, \traj, t)$ (or shortly as $r^\varphi$) is a real-valued function for signal $\traj$ and time $t$, with the value being positive iff $\traj \models \varphi$.
We recursively define $r$ for each operator as follows:
\begin{equation}
    \begin{aligned}
        r(\pi^\mu, \traj, t) =& \mu(\traj_t),\\
        r(\neg\varphi, \traj, t) =& -r(\pi^\mu, \traj, t) ,\\
        r(\varphi_1\land\varphi_2, \traj, t) =& \min (r(\varphi_1, \traj, t), r(\varphi_2, \traj, t)) ,\\
        r(\mathbf{F}_{[a,b]}\varphi, \traj, t) =& \max_{t_k \in [t+a, t+b]} (r(\varphi, \traj, t_k)) ,\\
        r(\varphi_1 \mathbf{U}_{[a,b]}\varphi_2, \traj, t) =& \max_{t_{k1} \in [t+a, t+b]} \big( \min (r(\varphi_1, \traj, t_{k1}), \min_{t_{k2} \in [t+a, t+t_{k1}]}r(\varphi_2, \traj, t_{k2}) ) \big).
    \end{aligned}
    \label{Eqn: SpaceRobustness}
\end{equation}

In this work, we utilize the modified robustness degree from \cite{varnai2020robustness} denoted as $\newRobust(\varphi_i, \traj, t)$ (or shortly as $\newRobust^{\varphi}$) because its properties are optimal for faster convergence \cite{dhonthi2022optimizing, dhonthi2021study}.
We define this robustness degree for the $\land$ operator as
\begin{equation}
    \begin{aligned}
        (\varphi_1\land\cdots\land\varphi_m) := 
        \begin{cases} 
            \dfrac{\sum_i r_{\min}e^{\newRobust_i}e^{\nu\newRobust_i}}{\sum_i e^{\nu\newRobust_i}} & \mathit   s{if} \; r_{\min} < 0, \\
            \dfrac{\sum_i r^{\varphi_i} e^{-\nu\newRobust_i}}{\sum_i e^{-\nu\newRobust_i}} & \mathit{if} \; r_{\min} > 0, \\
            0 & \mathit{if} \; r_{\min} = 0 ,
        \end{cases}
    \end{aligned}
    \label{Eqn: NewRob}
\end{equation}
with
\begin{equation}
\begin{aligned}
    r_{\min} = \min(r^{\varphi_i}\cdots r^{\varphi_m}), \
    \newRobust_i = \frac{r^{\varphi_i} - r_{\min}}{r_{\min}},
\end{aligned}
\end{equation}
where $\nu > 0$ is a hyper-parameter and tends to traditional space robustness as $\nu \rightarrow \infty$.

\vspace{-0.40cm}
\section{Methodology}
\label{sec: Methodology}
\vspace{-0.20cm}
We now utilize all the concepts defined above and introduce our method for optimizing model parameters for safe path planning.
After collecting the demonstrations and fitting an HSMM model, we obtain model parameters $\modelParams$ which represent the task at hand. 
Now, based on the real-world scenarios we identify safety properties, such as static and dynamic obstacles, task-specific safety properties, or road-rules and convert them to STL semantics $\varphi$.
We detail in Sec.~\ref{Sec: Safety properties}, how our approach converts the safety properties to STL and computes the robustness degrees.
We use these STL semantics and model parameters to optimize the parameters $\optimParams \subset \modelParams$.
At any optimization step $n \in N$, we reproduce the trajectory $\traj_n$ based on $\optimParams_n$, compute the robustness degree $\rho_n(\varphi, \traj_n, t)$ and, based on this, obtain new model parameters $\optimParams_{n+1}$.
Note that we initially modify the parameters $\optimParams_1$ randomly. 
At the end of $N$ steps, we obtain an optimal trajectory $\hat{\traj}$ that follows all the safety properties, given $\hat{\traj} \models \varphi$ and $\hat{\rho} > 0$.

The aforementioned method is suitable for a single instance of trajectory optimization.
However, automotive applications in real-world scenarios have longer time steps, and path planning for the whole trajectory at once is not feasible.
The reason is that the environment constantly changes, and the vehicle perception may be limited.
Therefore, we propose an adapted version called \emph{continuous multi-cycle path planning}.
We break the demonstrations to $M$ cycles, each cycle representing $T_m$ time steps.
This means that we divide the full task to $M$ sets giving rise to $M$ models, each represented as $\modelParams_m$.
At any cycle $m$, when the perception of the vehicle can cover the area of the next cycle $m+1$, we optimize $\optimParams_{m+1}$ based on the current perception.
Referring to the time taken to optimize the model parameters $\modelParams$ as $t^{\modelParams}$, our goal is to keep the time $t + t^{\modelParams}_{m+1} < T_m$, so that the vehicle motion is continuous from the initial state until the goal state.

\vspace{-0.40cm}
\subsection{Safety Properties as STL Specifications}
\vspace{-0.20cm}
\label{Sec: Safety properties}
In this section, we introduce our approach to convert the safety properties to logical specifications.
More specifically, the safety properties we define here are for avoiding static and dynamic obstacles, following traffic lights, and maintaining a safety distance to vehicles.
We first define the logical specification for the obstacles as
\begin{equation}
    \varphi_\mathit{obs} = \mathbf{G}_{[0,T]} \neg\varphi_{\mathit{obs}_1} \land \cdots \land \mathbf{G}_{[0,T]} \neg\varphi_{\mathit{obs}_O}
\end{equation}
with
\begin{equation}
\begin{aligned}
    \varphi_{\mathit{obs}_o} = (x_{o,lb} < x_o < x_{o,ub} )\; \land \; (y_{o,lb} < y_o < y_{o,ub}),
\end{aligned}
\end{equation}
where $x_o, y_o$ are the co-ordinate position of an obstacle $o$ coming from the observations, and we define the region of the obstacle with suffix $lb, ub$ representing the lower and upper bounds of the obstacle in $x$ and $y$ axis, respectively.
The robustness degree for this specification is defined as $\rho(\varphi_\mathit{obs}, \traj, t)$.

Dynamic obstacles, however, change their position at every time step $t$ and therefore, we define the obstacle positions as $x_o^t, y_o^t$ and their bounds as $x_{o,lb}^t, x_{o,lb}^t$.
We get the obstacle positions from the real-world scenario by identifying the direction and velocity of each obstacle. From that, we extract the positions and bounds at each time step, assuming that the obstacle continues to move in the same direction.
Our approach does not drastically affect the above-mentioned limitation because our total number of time-steps in one optimization cycle is small.
We can expand the predicate $\varphi_{\mathit{obs}_o}$ as
\begin{equation}
\begin{aligned}
    \varphi_{\mathit{obs}_o} = \varphi_{\mathit{obs}_o}^{t=1} \; \land \; \cdots \; \land \; \varphi_{\mathit{obs}_o}^{t=T}
\end{aligned}
\end{equation}
with
\begin{equation}
\begin{aligned}
    \label{Eqn: Dynamic inner predicate}
    \varphi_{\mathit{obs}_o}^{t} = (x_{o,lb}^t < x_o^t < x_{o,ub}^t )\; \land \; (y_{o,lb}^t < y_o^t < y_{o,ub}^t).
\end{aligned}
\end{equation}
The computation of the robustness degree $\rho(\varphi_\mathit{obs}, \traj, t)$ remains the same because we can directly use Eq~\ref{Eqn: NewRob} due to the $\land$ operators between each predicate $\varphi_{\mathit{obs}_o}^{t}$.
The only difference is that the inner predicates defined in Eq~\ref{Eqn: Dynamic inner predicate} change at each time step $t$.

Similarly, we can define the STL specification for the road rules.
The road rules can be of various kinds, for example, we can define the rule not to cross into the opposite lane by simply setting the opposite lane as a static obstacle.
Some complex properties such as staying behind a traffic light until it is green can be formulated using the until operator as
\begin{equation}
    \varphi_\mathit{safe} = \varphi_\mathit{avoid} \mathbf{U}_{[t_1, t_2]} \varphi_\mathit{stay},
\end{equation}
where the definition of $\varphi_\mathit{avoid}$ is similar to the constraint for static obstacles, so that the region at the cross roads is avoided. $t_1, t_2$ define the time during which the traffic light stays red and $\varphi_\mathit{stay}$ defines the event that the traffic light turns green.

\vspace{-0.30cm}
\section{Experiments}
\vspace{-0.20cm}

\begin{figure}[t]
    \centering
	\includegraphics[width=7.0cm]{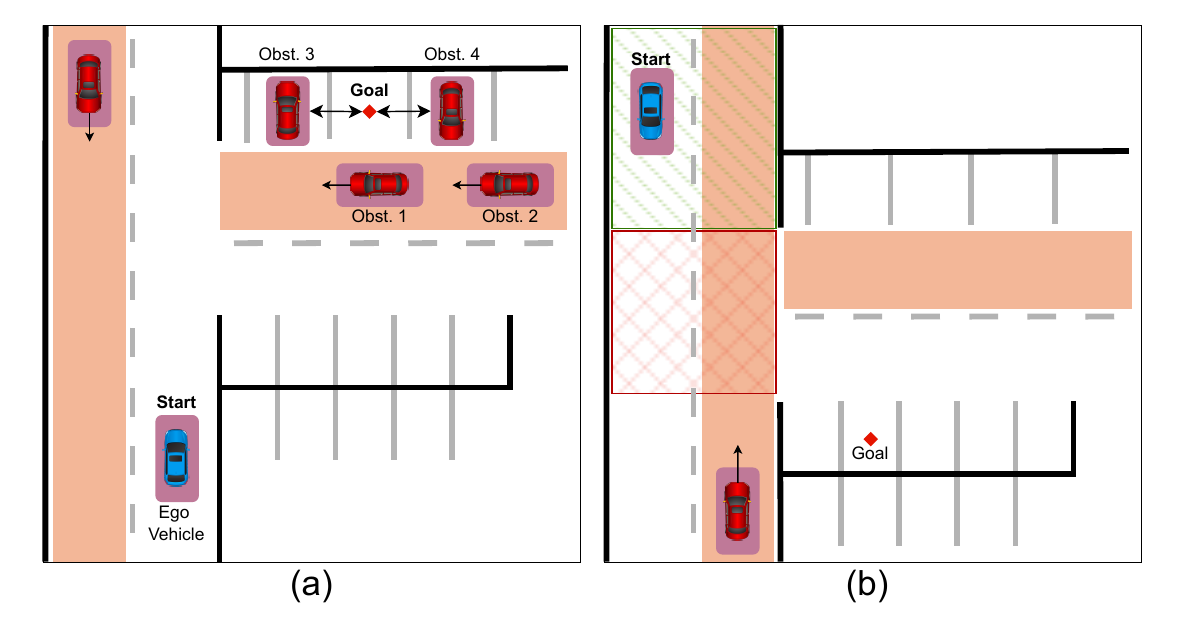}
	\caption{Valet parking scenarios. The ego vehicle (in blue) must move from start to goal (depicted as diamond) while avoiding static (in red) and dynamic (in red with an arrow) obstacles. In (a), we depict the safety distance between the adjacent vehicles and the goal state. In (b), we depict the region to avoid (red cross-hatched) and region to stay (green hatched) when traffic light is red.}
	\label{Fig: Scenarios}
	\vspace{-0.20cm}
\end{figure}

We utilize the \emph{IR-SIM} simulation environment~\cite{IRSIM2024} to define two real-time automated valet parking scenarios.
In these scenarios, the vehicle must plan a trajectory to reach a goal state which is a pre-defined parking place.
We depict the two scenarios in Fig.~\ref{Fig: Scenarios} which consist of static and dynamic obstacles, some safety restrictions and a traffic light at the junction.
For each scenario, we record $4$ trajectories $\traj_{i=1}^{N=4}$ with each $\traj$ lasting for $T=20$ seconds.
The value of $\nu$ is set to $5.0$ for optimal results based on multiple experimental evaluations.
We use these scenarios to evaluate our single-cycle path planning and continuous multi-cycle path planning algorithms.
The evaluation is based on the ability to address all the set constraints, and based on the time taken to obtain optimal trajectories. 

\begin{figure}[t]
	\includegraphics[width=\textwidth, trim={5.0cm 0.0cm 5.0cm 0.0cm},clip]{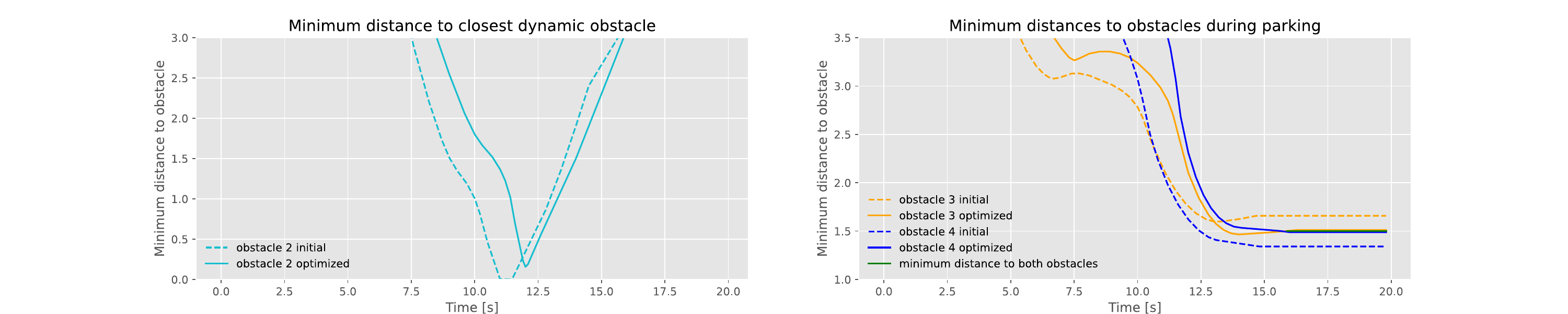}
	\caption{We depict the distance to the obstacles over time before optimization (in dotted lines) and after optimization (in solid lines). A collision occurs when the distance is $0.0$. The minimum distance constraint (on the left) during time $16$ to $20$ seconds is depicted as green line.}
	\label{Fig: Collision}
	\vspace{-0.50cm}
\end{figure}

The goal of Scenario (a) is to avoid the obstacles and to maintain a minimum safety distance to the vehicles adjacent to the parking place.
We define the STL specification for the first scenario as
\begin{equation}
    \varphi_1 = \mathbf{G}_{[0,20]} \neg\varphi_{\mathit{obs}_O} \land \mathbf{G}_{[0,20]} \varphi_{\mathit{rules}} \land \mathbf{F}_{[16,20]} \varphi_{\mathit{safe}},
\end{equation}
where $\varphi_{\mathit{obs}_O}$ are the $5$ obstacles as depicted in Fig.~\ref{Fig: Scenarios}(a).
The road rule to not cross to the opposite lane is defined as a region $\varphi_{\mathit{rules}}$.
We define the safety distance between parked vehicles and ego vehicle as $\varphi_{\mathit{safe}} = x_o^t - x_{\mathit{ego}}^t < 1.5$, where $x_{\mathit{ego}}^t$ is the position of the ego vehicle at time $t$ in $x-axis$.
Note that the time frame 16 to 20 seconds is identified from the simulation. 
In a continuous planner, we obtain the correct time intervals for the logic in real-time when the ego vehicle is close to the two vehicles and accordingly the trajectory in the next cycle is optimized. 
We can also relax the time interval restriction for $\varphi_{\mathit{rules}}$ to a specific time if the car has to use other lanes, for instance when it has to cross the opposite lane for parking.

Fig.~\ref{Fig: Collision} depicts the results of Scenario (a). 
As we can see, the collision into the dynamic obstacle $\varphi_{\mathit{obs}_2}$ is avoided after the optimization.
Also, the figure on the right shows that the minimum distance to the parked vehicles is also achieved as the distance to both obstacles coincides at $1.5$ m.
Overall, we achieved a positive reward, which means that the optimized trajectory addresses all the constraints set in $\varphi_1$.

\begin{figure}[t]
    \centering
	\includegraphics[width=7.5cm]{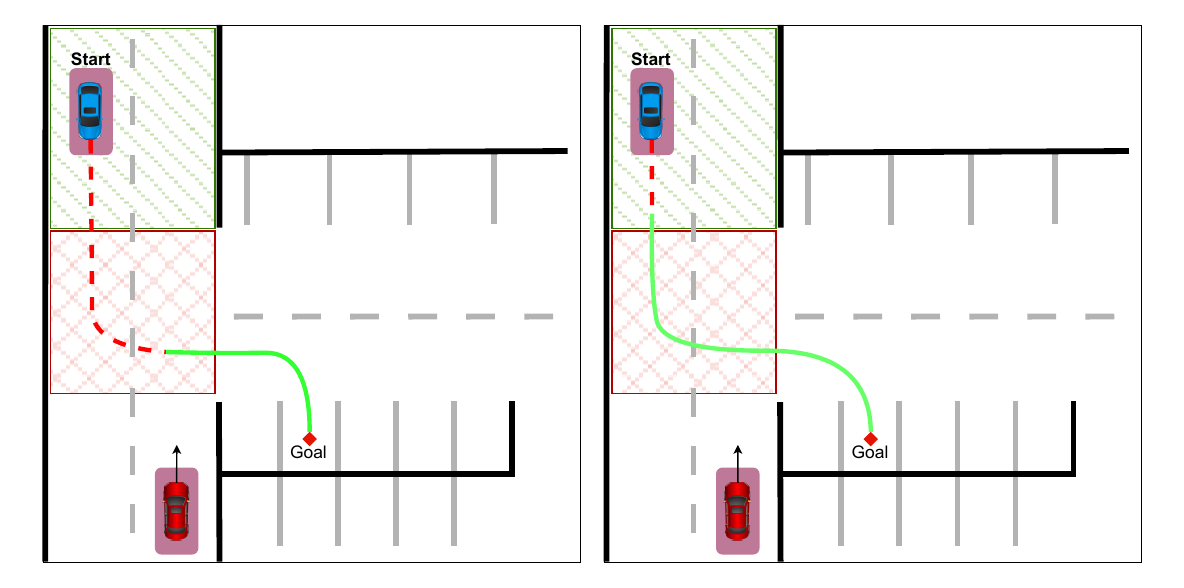}
	\caption{We depict the ego vehicle trajectory when traffic light is red (as dotted red line) and when it is green (as solid green line) for both before optimization (left) and after optimization (right).}
	\label{Fig: Traffic Scenario}
\end{figure}

Similarly, the goal of Scenario (b) is to avoid the obstacles and stay behind the junction in the first 4 seconds, when the traffic light is red.
The STL specification for this scenario is
\begin{equation}
    \varphi_2 = \mathbf{G}_{[0,20]} \neg\varphi_{\mathit{obs}_O} \land \mathbf{G}_{[0,20]} \varphi_{\mathit{rules}} \land \varphi_\mathit{avoid} \mathbf{U}_{[0, 4]} \varphi_\mathit{stay},
\end{equation}
where, as depicted in Fig.~\ref{Fig: Scenarios}(b), $\varphi_\mathit{avoid}$ is the red cross-hatched region and $\varphi_\mathit{stay}$ is the green hatched region. 
Before optimization, Fig.~\ref{Fig: Traffic Scenario} depicts the ego vehicle moving into the junction when the traffic light is red. This is avoided in the optimized trajectory because the ego vehicle waits behind the junction until the traffic light is green.

Table~\ref{tab:runtime} depicts the optimization results of Scenario (a) for the continuous planner with minimal observation divided into $4$ cycles. 
We partition the whole task into $M=4$ cycles and at each cycle, we have the observation of obstacles in the next cycle. 
As we can see in Table~\ref{tab:runtime}, we obtain positive rewards in every cycle, which means all the constraints were satisfied. 
Additionally, as mentioned in Sec.~\ref{sec: Methodology}, the time taken for optimization must be less than the total time of that cycle.
We also achieved this, as depicted in the last two columns of the table.
Therefore, we can say that our algorithm also works for continuous local planning when the observation is minimal.

\begin{table}[t]
\centering
\caption{Runtime measurements for continuous multi-cycle path planning}
{
\setlength{\tabcolsep}{2pt}
\begin{tabular}{c|c|c|c|c}
Cycle & Initial robustness & Optimized robustness & Optimization time [s] & Simulation time [s]\\
\hline
1 & - & - & 6.73 & -\\
2 & 1.432 & 1.619 & 7.118 & 7.65\\
3 & 1.199 & 1.278 & 6.636 & 8.172\\
4 & -0.045 & 0.091 & 5.887 & 9.412\\
5 & 0.016 & 0.035 & - & 12.118\\
\hline
\end{tabular}
}
\label{tab:runtime}
\end{table}

The results above show that our method is powerful to incorporate various types of constraints and to address all of them at once to achieve safe trajectories.
Using our reward function, we can verify that the generated trajectories are safe during the runtime. 
Our optimization algorithm can work for both minimal and full observation.

\vspace{-0.30cm}
\section{Conclusion}
\vspace{-0.30cm}
In this paper, we addressed the verification and optimization of path planning trajectories during runtime with both minimal and full observations.
We defined static and dynamic obstacles, along with constraints from safety standards in the form of STL specifications, and used it to obtain optimal trajectories.
Future work would include testing the approach on critical real-world scenarios and incorporating complex constraints using STL.

%
%
%
\bibliographystyle{splncs04}
\bibliography{Bibliography}

\end{document}